\documentclass[12pt]{spieman}  
\usepackage{amsmath,amsfonts,amssymb}
\usepackage{graphicx}
\usepackage{setspace}
\usepackage{tocloft}
\usepackage{algorithm}
\usepackage{algorithmic}
\usepackage{lineno}
\newcommand{\etal}{\textit{et al.}}
\newcommand{\ie}{i.e.}

\title{Class-specific residual constraint non-negative representation for pattern classification}

\author[a,b]{He-Feng Yin}
\author[a,b,*]{Xiao-Jun Wu}
\affil[a]{Jiangnan University, School of Internet of Things Engineering, No. 1800, Lihu Avenue, Wuxi, China, 214122}
\affil[b]{Jiangsu Provincial Laboratory of Pattern Recognition and Computational Intelligence, No. 1800, Lihu Avenue, Wuxi, China, 214122}

\cftpagenumbersoff{figure}
\cftpagenumbersoff{table} 
\begin{document} 
\maketitle

\begin{abstract}
Representation based classification method (RBCM) remains one of the hottest topics in the community of pattern recognition, and the recently proposed non-negative representation based classification (NRC) achieved impressive recognition results in {\color{red}various classification tasks}. However, {\color{red}NRC ignores the relationship between the coding and classification stages. Moreover,} there is no regularization term other than the reconstruction error term in the formulation of NRC, which may result in unstable solution leading to misclassification. To overcome {\color{red}these drawbacks} of NRC, in this paper, we propose a class-specific residual constraint non-negative representation (CRNR) for pattern classification. CRNR introduces a class-specific residual constraint into the formulation of NRC, which encourages {\color{red}training samples from different classes to competitively represent the test sample.} Based on the proposed CRNR, we develop a CRNR based classifier (CRNRC) for pattern classification. Experimental results on several benchmark datasets demonstrate the superiority of CRNRC over conventional RBCM as well as the recently proposed NRC. Moreover, CRNRC works better or comparable to some state-of-the-art deep approaches on diverse challenging pattern classification tasks. The source code of our proposed CRNRC is accessible at \url{https://github.com/yinhefeng/CRNRC}.
\end{abstract}


{\noindent \footnotesize\textbf{*}Xiao-Jun Wu,  \linkable{wu\_xiaojun@jiangnan.edu.cn} }

\begin{spacing}{2}   

\section{Introduction}
\label{sect:intro}  
During the past few years, representation based classification method (RBCM) has received considerable attention in various classification tasks, such as face classification~\cite{song2019collaborative} and hyperspectral image classification~\cite{jia2019collaborative}. In the domain face recognition, the popular work is the sparse representation based classification (SRC)~\cite{wright2009robust}. SRC treats all the training samples as a dictionary, and a test sample is sparsely coded over the dictionary {\color{red}with the $\ell_1$-norm constraint}, then the classification is accomplished by checking which class yields the least reconstruction error. SRC can achieve excellent recognition results even when the test samples are occluded or corrupted. {\color{red}Due to the principle of $\ell_1$-minimization, SRC tends to lose locality information. However, Yu \etal~\cite{yu2009nonlinear} revealed that under certain assumptions locality is more essential than sparsity. To incorporate the locality structure of data into the sparse coding of test data, Wang \etal~\cite{wang2010locality} developed a locality-constraint linear coding (LLC) scheme. Similarly, Lu \etal~\cite{lu2013face} proposed a weighted sparse representation based classification (WSRC) method. Different from the $\ell_1$-norm used in WSRC, LLC employs $\ell_2$-norm in which analytical solution can be derived. When the size of the dictionary is huge, the $\ell_1$-norm minimization problem in SRC is computationally expensive. To speed up the sparse coding process, Li \etal~\cite{li2010local} presented a local sparse representation based classification (LSRC) scheme which performs sparse decomposition in local neighborhood. Ortiz \etal~\cite{ortiz2014face} designed a linearly approximated sparse representation-based classification (LASRC) algorithm that employs linear regression to perform sample selection for $\ell_1$-minimization.}

Another prevailing approach of RBCM is collaborative representation based classification (CRC). Zhang \etal~\cite{zhang2011sparse} argued that it is the collaborative representation (CR) mechanism rather than the $\ell_1$-norm sparsity that makes SRC powerful for classification. {\color{red}Compared with SRC, CRC can achieve comparable performance but the computational complexity is significantly reduced.} Similarly, Xu \etal~\cite{xu2016new} introduced a discriminative sparse representation (DSR) method for robust face recognition via $\ell_2$ regularization. {\color{red}Timofte \etal~\cite{timofte2012weighted} investigated the weighted variants of collaborative representation and proposed weighted collaborative representation (WCR) for image classification. To fully exploit the strengths of diverse RBCMs, Chi \etal~\cite{chi2013classification} presented a collaborative representation optimized classifier (CROC) which achieves a balance between the nearest subspace classifier (NSC)~\cite{lee2005acquiring} and CRC.} Although Zhang \etal~\cite{zhang2011sparse} offered a geometric interpretation of the classification mechanism of CRC, it is still obscure to understand its intrinsic principle. Afterwards, Cai \etal~\cite{cai2016probabilistic} analyzed the classification mechanism of CRC from a probabilistic viewpoint and proposed a probabilistic collaborative representation based classifier (ProCRC). Based on ProCRC, Gou \etal~\cite{gou2019two} presented two-phase probabilistic collaborative representation based-classification (TPCRC) which adopts the coarse to fine strategy.

Although SRC and CRC (and their extensions) achieve impressive recognition results in various classification tasks, they cannot avoid negative entries in their coding vectors for test samples. The negative coefficients indicate negative data correlations between the test sample and the training samples.  Inspired by the principle of non-negative matrix factorization (NMF)~\cite{lee1999learning}, Xu \etal~\cite{xu2019sparse} proposed a non-negative representation based classifier (NRC) which introduces a non-negative constraint on the coding vector. Extensive experiments on diverse classification tasks demonstrate the superiority of NRC over many existing RBCM, including SRC, CRC and ProCRC. {\color{red}NRC consists of two separate stages: firstly, employing all the training samples to represent the test sample; secondly, classifying the test sample into the class that yields the least residual.} Nevertheless, besides the reconstruction error term, there is no other regularization terms in the formulation of NRC, which may produce unstable solution and lead to misclassification. {\color{red}Moreover, NRC ignores the relationship between the first and second stage.} To tackle {\color{red}these problems}, we incorporate a class-specific residual constraint into the formulation of NRC and propose a class-specific residual constraint non-negative representation (CRNR) for pattern classification. Through this constraint, {\color{red}training samples across different classes are encouraged to competitively reconstruct the test sample. We find that minimizing this constraint is equivalent to the NSC model. Therefore, CRNR can be regarded as a combination of the CRC and NSC models with the non-negative constraint to obtain the coding vector of test sample.}

\begin{figure}[htbp]
  \centering
  \includegraphics[trim={0mm 0mm 0mm 0mm},clip, width = .8\textwidth]{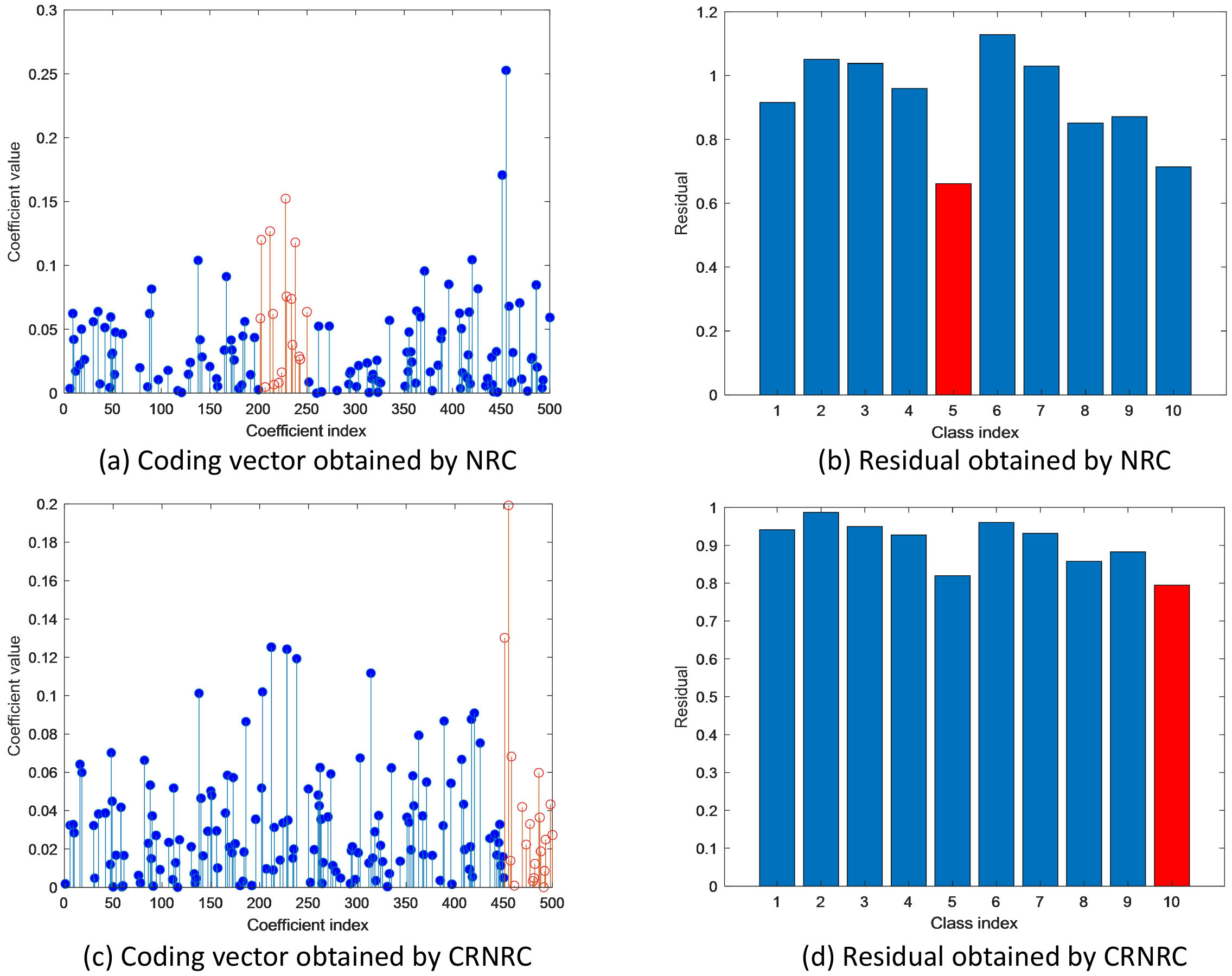}
  \caption{Coding vector and residual obtained by NRC and CRNRC, respectively. The test sample belongs to the tenth class, i.e., digit 9. (a) Coding vector obtained by NRC; (b) Class-specific residual of NRC, one can see that the fifth class yields the least residual, thus the test sample digit 9 is misclassified into digit 4; (c) Coding vector obtained by our proposed method; (d) Class-specific residual obtained by CRNRC, we can observe that the tenth class leads to the minimum residual. Therefore, the test sample is correctly classified into digit 9.}
  \label{fig:coeff_res}
\end{figure}

To illustrate the mechanism of CRNRC, we carry out an experiment on the MNIST dataset. This dataset contains images for digits 0-9, and 50 images per class are selected to form the training set. The 500 images are arranged in an order of $[0,1,2,\ldots,9]$, thus the training data matrix is denoted by $\mathbf{X}=[\mathbf{X}_0,\mathbf{X}_1,\ldots,\mathbf{X}_9]$, and we choose a test sample from the tenth class (\ie, digit 9). The coding vector and residual obtained by NRC are shown in Figs.~\ref{fig:coeff_res} (a) and (b), respectively, while the coding vector and residual obtained by CRNRC are shown in Figs.~\ref{fig:coeff_res} (c) and (d), respectively. From Fig.~\ref{fig:coeff_res} (b), one can see that the fifth class has the least residual, \ie, the test sample is recognized as digit 4 even though the tenth class has dominant coefficients in Fig.~\ref{fig:coeff_res} (a). From Fig.~\ref{fig:coeff_res} (d), we can observe that the tenth class results in the minimum residual, \ie, the test sample is correctly recognized as digit 9. When we have a closer look at the coding vector of NRC and CRNRC, 
for the fifth class (indices 201-250), the number of nonzero entries of NRC and CRNRC are 16 and 14, respectively. While for the tenth class (indices 451-500), the number of nonzero entries of NRC and CRNRC are 15 and 19, respectively. By introducing the class-specific residual constraint into NRC, more coefficients are concentrated on the correct class, thus improved performance of our proposed CRNRC can be expected.

Our main contributions are summarized as follows,
\begin{itemize}
\item A class-specific residual constraint non-negative representation (CRNR) scheme is proposed. Moreover, a CRNR based classifier (CRNRC) is designed for pattern classification.
\item The resulting optimization problem of CRNR is elegantly solved under the framework of alternating direction method of multipliers (ADMM)~\cite{boyd2011distributed}.
\item Our proposed CRNRC outperforms conventional RBCM and works better or comparable to some state-of-the-art deep approaches on various standard datasets.
\end{itemize}

The rest of this paper is structured as follows: Section \ref{sect:sect_2} reviews the related work. Section \ref{sect:sect_3} presents our CRNRC approach. Experimental evaluation on several benchmark databases is presented in Section \ref{sect:sect_4}. Finally, Section \ref{sect:sect_5} concludes this paper.

\section{Related Work}
\label{sect:sect_2}
In this section, we will briefly review some related work, including SRC~\cite{wright2009robust}, CRC~\cite{zhang2011sparse} and NRC~\cite{xu2019sparse}. To begin with, we present some notations used throughout this paper. Suppose we have $n$ training samples from $K$ classes, and the training data matrix is denoted by $\mathbf{X}=\left[\mathbf{X}_{1}, \mathbf{X}_{2}, \ldots, \mathbf{X}_{K}\right]=[\boldsymbol{x}_1,\boldsymbol{x}_2,\ldots,\boldsymbol{x}_n] \in \mathbb{R}^{d \times n}$, where $\mathbf{X}_i$ is the data matrix of the $i$-th class. The $i$-th class has $n_i$ training samples and $\sum_{i=1}^Kn_i=n$ ($i=1,2,\ldots,K$), $d$ is the dimensionality of vectorized samples.

\subsection{Sparse Representation based Classification}
\label{sect:title}
In SRC, a test sample $\boldsymbol{y}\in\mathbb{R}^d$ is firstly expressed as a sparse linear superposition of all the training data, then the classification is done by checking which class yields the minimum reconstruction error, the objective function of SRC can be formulated as,
\begin{equation}
\label{eq:obj_src}
\underset{\boldsymbol{c}}{\textrm{min}} \ \left \| \boldsymbol{y}-\mathbf{X}\boldsymbol{c} \right \|_2^2+\lambda\left \| \boldsymbol{c} \right \|_1
\end{equation}
where $\lambda>0$ is a balancing parameter. Once we obtain the coefficient vector $\boldsymbol{c}$ of $\boldsymbol{y}$, the test sample $\boldsymbol{y}$ is classified according to the following formulation,
\begin{equation}
\label{eq:rule_src}
\textrm{identity}\left( \boldsymbol{y} \right)=\arg \underset{i}{\mathop{\min }}\,\left\| \boldsymbol{y}-{{\mathbf{X}}_{i}}{{\boldsymbol{c} }_{i}} \right\|_{2}
\end{equation}
where ${\boldsymbol{c} }_{i}$ is the coefficient vector that corresponds to the $i$-th class.

\subsection{Collaborative Representation based Classification}
Zhang \etal~\cite{zhang2011sparse} presented collaborative representation based classification (CRC) algorithm, which replaces the $\ell_1$-norm in SRC with the $\ell_2$-norm constraint, the objective function of CRC is formulated as follows,
\begin{equation}
\label{eq:obj_crc}
\underset{\boldsymbol{c}}{\textrm{min}} \ \left \| \boldsymbol{y}-\mathbf{X}\boldsymbol{c} \right \|_2^2+\lambda\left \| \boldsymbol{c} \right \|_2^2
\end{equation}
CRC has the following closed-form solution,
\begin{equation}
\label{eq:solu_crc}
\boldsymbol{c}=(\mathbf{X}^T\mathbf{X}+\lambda \mathbf{I})^{-1}\mathbf{X}^T\boldsymbol{y}
\end{equation}
where $\mathbf{I}$ is an identity matrix. Let $\mathbf{P}=(\mathbf{X}^T\mathbf{X}+\lambda \mathbf{I})^{-1}\mathbf{X}^T$, we can see that $\mathbf{P}$ is only determined by the training data matrix $\mathbf{X}$. Therefore, when given all the training data, $\mathbf{P}$ can be pre-computed, which makes CRC very efficient. CRC employs the following regularized residual for classification,
\begin{equation}
\label{rule_crc}
\textrm{identity}\left( \boldsymbol{y} \right)=\arg \underset{i}{\mathop{\min }}\,\frac{\left\| \boldsymbol{y}-{{\mathbf{X}}_{i}}{{\boldsymbol{c} }_{i}} \right\|_{2}}{\left \| \boldsymbol{c}_i \right \|_2} 
\end{equation}

\subsection{Nonnegative Representation based Classification}
SRC and CRC have become two representative approaches in RBCM. However, the coding vector of conventional RBCM contains negative entries. The test sample should be better expressed by homogeneous samples with non-negative representation coefficients. Moreover, Lee and Seung~\cite{lee1999learning} pointed out that it is unsuitable to approximate the test sample by allowing the training samples to “cancel each other out” with complex additions and subtractions. Therefore, Xu \etal~\cite{xu2019sparse} proposed the following non-negative representation model by imposing the non-negative constraint on the coding vector, 
\begin{equation}
\label{eq:obj_nrc}
\underset{\boldsymbol{c}}{\textrm{min}} \ \left \| \boldsymbol{y}-\mathbf{X}\boldsymbol{c} \right \|_2^2, \ \textrm{s.t.} \ \boldsymbol{c}\geq 0
\end{equation}
Similar to SRC, NRC employs the class specific residual to classify the test sample, \ie, $\textrm{identity}\left( \boldsymbol{y} \right)=\arg \underset{i}{\mathop{\min }}\,\left\| \boldsymbol{y}-{{\mathbf{X}}_{i}}{{\boldsymbol{c} }_{i}} \right\|_{2}$.

\section{Class-specific Residual constraint Nonnegative Representation}
\label{sect:sect_3}
In this section, first we present the model of our proposed CRNR, then we design a CRNR based classifier (CRNRC) for pattern classification.
\subsection{Proposed Model}
From Equation~(\ref{eq:obj_nrc}), we can see that apart from the reconstruction error term, there is no other regularized terms in the objective function of NRC. As demonstrated by the illustration in Section~\ref{sect:intro}, due to lack of regularization, NRC would result in misclassification. {\color{red}Furthermore, NRC ignores the relationship between the coding and classification stages.} To alleviate {\color{red}these problems}, we incorporate a class-specific residual constraint into the formulation of NRC, and the objective function of our CRNR is formulated as follows,
\begin{equation}
\label{eq:obj_crnr}
\underset{\boldsymbol{c}}{\textrm{min}} \ \left \| \boldsymbol{y}-\mathbf{X}\boldsymbol{c} \right \|_2^2+\lambda\sum_{i=1}^{K}\left \| \boldsymbol{y}-\mathbf{X}_i\boldsymbol{c}_i \right \|_2^2, \ \textrm{s.t.} \ \boldsymbol{c}\geq 0
\end{equation}
where $\lambda>0$ is a balancing parameter. The first term in Eq.~(\ref{eq:obj_crnr}) is the collaborative representation term and the second term is the class-specific residual constraint. {\color{red}For a specific class, the second term becomes $\left \| \boldsymbol{y}-\mathbf{X}_i\boldsymbol{c}_i \right \|_2^2$, which is the formulation of NSC~\cite{lee2005acquiring}. Therefore, our proposed CRNR can be viewed as an integration of CRC and NSC with the non-negative constraint to compute the representation.} One can see that when $\lambda = 0$, CRNR is degenerated to NRC. Therefore, NRC can be viewed as a special version of CRNRC.

\subsection{Optimization}
We adopt an alternative strategy to solve the CRNR model. By introducing an auxiliary variable $\boldsymbol{z}$, Equation~(\ref{eq:obj_crnr}) can be rewritten as,
\begin{equation}
\label{eq:obj_equi}
\underset{\boldsymbol{c},\boldsymbol{z}}{\textrm{min}} \ \left \| \boldsymbol{y}-\mathbf{X}\boldsymbol{c} \right \|_2^2+\lambda\sum_{i=1}^{K}\left \| \boldsymbol{y}-\mathbf{X}_i\boldsymbol{c}_i \right \|_2^2, \ \textrm{s.t.} \ \boldsymbol{c}=\boldsymbol{z},\boldsymbol{z}\geq 0
\end{equation} 

Equation~(\ref{eq:obj_equi}) can be solved by the alternating direction method of multipliers (ADMM)~\cite{boyd2011distributed} technique, and the Lagrangian function of Eq.~(\ref{eq:obj_equi}) is,
\begin{equation}
\label{eq:obj_lagrange}
\mathcal{L}(\boldsymbol{c},\boldsymbol{z},\boldsymbol{\delta},\mu)= \left \| \boldsymbol{y}-\mathbf{X}\boldsymbol{c} \right \|_2^2+\lambda\sum_{i=1}^{K}\left \| \boldsymbol{y}-\mathbf{X}_i\boldsymbol{c}_i \right \|_2^2+\left \langle \boldsymbol{\delta},\boldsymbol{z}-\boldsymbol{c} \right \rangle+\frac{\mu}{2}\left \| \boldsymbol{z}-\boldsymbol{c} \right \|_2^2
\end{equation}
where $\boldsymbol{\delta}$ is the Lagrange multiplier and $\mu>0$ is a penalty parameter. The optimization of Eq.~(\ref{eq:obj_lagrange}) can be solved iteratively by updating $\boldsymbol{c}$ and $\boldsymbol{z}$ once at a time. The detailed updating procedures are presented as follows.

\textit{Update} $\boldsymbol{c}$: Fix the other variables and update $\boldsymbol{c}$ by solving the following problem,

\begin{equation}
\label{eq:update_c}
\underset{\boldsymbol{c}}{\textrm{min}} \ \left \| \boldsymbol{y}-\mathbf{X}\boldsymbol{c} \right \|_2^2+\lambda\sum_{i=1}^{K}\left \| \boldsymbol{y}-\mathbf{X}_i\boldsymbol{c}_i \right \|_2^2+\frac{\mu}{2}\left \| \boldsymbol{z}_t-\boldsymbol{c} +\frac{\boldsymbol{\delta}_t}{\mu}\right \|_2^2
\end{equation}
Suppose $\mathbf{X}_i^{'}$ is a matrix that has the same size as $\mathbf{X}$, and $\mathbf{X}_i^{'}$ only consists of samples from the $i$-th class, \ie, $\mathbf{X}_i^{'}=[\boldsymbol{0},\ldots,\mathbf{X}_i,\ldots,\boldsymbol{0}]$, Equation~(\ref{eq:update_c}) can be reformulated as,
\begin{equation}
\label{eq:update_c_equi}
\underset{\boldsymbol{c}}{\textrm{min}} \ \left \| \boldsymbol{y}-\mathbf{X}\boldsymbol{c} \right \|_2^2+\lambda\sum_{i=1}^{K}\left \| \boldsymbol{y}-\mathbf{X}_i^{'}\boldsymbol{c} \right \|_2^2+\frac{\mu}{2}\left \| \boldsymbol{z}_t-\boldsymbol{c} +\frac{\boldsymbol{\delta}_t}{\mu}\right \|_2^2
\end{equation}
Setting the partial derivative of Eq.~(\ref{eq:update_c_equi}) with respect to $\boldsymbol{c}$ to zero, we can obtain the following closed-form solution,
\begin{equation}
\label{eq:solu_c}
\boldsymbol{c}_{t+1}=(\mathbf{X}^T\mathbf{X}+\lambda \sum_{i=1}^{K}(\mathbf{X}_i^{'})^T(\mathbf{X}_i^{'})+\frac{\mu}{2}\mathbf{I})^{-1}[(1+\lambda)\mathbf{X}^T\boldsymbol{y}+\frac{\mu \boldsymbol{z}_t+\boldsymbol{\delta}_t}{2}]
\end{equation}

\textit{Update} $\boldsymbol{z}$: To update $\boldsymbol{z}$, we fix variables other than $\boldsymbol{z}$ and solve the following problem accordingly,
\begin{equation}
\label{eq:update_z}
\underset{\boldsymbol{z}}{\textrm{min}} \ \left \| \boldsymbol{z}-(\boldsymbol{c}_{t+1}-\frac{\boldsymbol{\delta}_t}{\mu}) \right \|_2^2, \ \textrm{s.t.} \ \boldsymbol{z}\geq 0
\end{equation}
The solution to $\boldsymbol{z}$ is given by,
\begin{equation}
\label{eq:solu_z}
\boldsymbol{z}_{t+1}=\textrm{max}(0,\boldsymbol{c}_{t+1}-\frac{\boldsymbol{\delta}_t}{\mu})
\end{equation}
where the “max” operator performs element by element.

\textit{Update} $\boldsymbol{\delta}$: 
The Lagrange multiplier $\boldsymbol{\delta}$ is updated according to the following formulation,
\begin{equation}
\label{eq:update_delta}
\boldsymbol{\delta}_{t+1}=\boldsymbol{\delta}_t+\mu(\boldsymbol{z}_{t+1}-\boldsymbol{c}_{t+1})
\end{equation}
The detailed procedures of solving the Eq.~(\ref{eq:obj_crnr}) are described in Algorithm~\ref{alg1}.

\begin{algorithm} 
\caption{Solve Eq.~(\ref{eq:obj_crnr}) via ADMM} 
\label{alg1} 
\begin{algorithmic}[1]
\REQUIRE Test sample $\boldsymbol{y}$, training data matrix $\mathbf{X}$, balancing parameter $\lambda$, $\textrm{tol}>0$, $\mu>0$ and the maximum iteration number $T$.
\STATE Initialize $\boldsymbol{z}_0=\boldsymbol{c}_0=\boldsymbol{\delta}_0=\boldsymbol{0}$;
\WHILE{not converged} 
\STATE Update $\boldsymbol{c}$ by Eq.~(\ref{eq:solu_c});
\STATE Update $\boldsymbol{z}$ by Eq.~(\ref{eq:solu_z});
\STATE Update $\boldsymbol{\delta}$ by Eq.~(\ref{eq:update_delta});
\ENDWHILE 
\ENSURE Coding vectors $\boldsymbol{z}$ and $\boldsymbol{c}$.
\end{algorithmic} 
\end{algorithm}

\subsection{Classification}
For the test sample $\boldsymbol{y}\in\mathbb{R}^d$, first we obtain its coding vector $\boldsymbol{c}$ over the entire training data $\mathbf{X}$ by solving Eq.~(\ref{eq:obj_crnr}), then the test sample is designated into the class that yields the least residual, \ie, $\textrm{identity}(\boldsymbol{y})=\textrm{arg} \ \underset{i}{\textrm{min}}\left \| \boldsymbol{y}-\mathbf{X}_i\boldsymbol{c}_i \right \|_2$, where $\boldsymbol{c}_i$ is the coding vector that belongs to the $i$-th class. The complete process of our proposed CRNRC is summarized in Algorithm~\ref{alg2}.
\begin{algorithm}[t]
\begin{algorithmic}[1]
\vspace{0.03in}
\REQUIRE Training data matrix $\mathbf{X}=\left[\mathbf{X}_{1}, \mathbf{X}_{2}, \ldots, \mathbf{X}_{K}\right] \in \mathbb{R}^{d \times n}$, test data $\boldsymbol{y} \in \mathbb{R}^{d}$ and balancing parameter $\lambda$.
\STATE Normalize the columns of $\mathbf{X}$ and $\boldsymbol{y}$ to have unit $\ell_2$ norm;
\STATE Obtain the coding vector $\boldsymbol{c}$ of $\boldsymbol{y}$ on $\mathbf{X}$ by solving the CRNR model in Eq.~(\ref{eq:obj_crnr});
\STATE Compute the class-specific residuals $\boldsymbol{r}_i=\left \| \boldsymbol{y}-\mathbf{X}_i\boldsymbol{c}_i \right \|_2$;
\ENSURE $\textrm{label}(\boldsymbol{y})=\arg \min_{i}\left(\boldsymbol{r}_{i}\right)$
\vspace{0.03in}
\end{algorithmic}
\caption{Our proposed CRNRC algorithm}
\label{alg2}
\end{algorithm}
\section{Experiments}
\label{sect:sect_4}
In this section, we evaluate the classification performance of CRNRC on diverse benchmark datsets: {\color{red}three} face databases including AR~\cite{martinez1998ar}, Extended Yale B~\cite{georghiades2001few} {\color{red}and Georgia Tech (GT) ~\cite{feng2016superimposed}} databases, two handwritten digit datasets including USPS~\cite{hull1994database} and MNIST~\cite{lecun1998gradient} datasets, and four large-scale datasets including the Stanford 40 dataset~\cite{yao2011human}, the Oxford 102 Flowers dataset~\cite{nilsback2008automated}, the Aircraft dataset~\cite{maji2013fine} and the Cars dataset~\cite{krause20133d}. We compare the classification accuracy of CRNRC with NSC~\cite{lee2005acquiring}, linear SVM, SRC~\cite{wright2009robust}, CRC~\cite{zhang2011sparse}, CROC~\cite{chi2013classification}, ProCRC~\cite{cai2016probabilistic} and NRC~\cite{xu2019sparse}. In addition, {\color{red}on the GT database, we compare CRNRC with several inception deep architectures, \ie, GoogLeNet~\cite{szegedy2015going}, Inception-v3~\cite{szegedy2016rethinking}, Xception~\cite{chollet2017xception} and Inception-ResNet-v2~\cite{szegedy2017inception}. On the four large-scale datasets,} we compare CRNRC with the state-of-the-art (SOTA) methods.
\subsection{Face Recognition}
\subsubsection{Experiments on the AR Database}
\label{sec:4_1_1}

\begin{figure}[htbp]
  \centering
  \includegraphics[trim={0mm 0mm 0mm 0mm},clip, width = .8\textwidth]{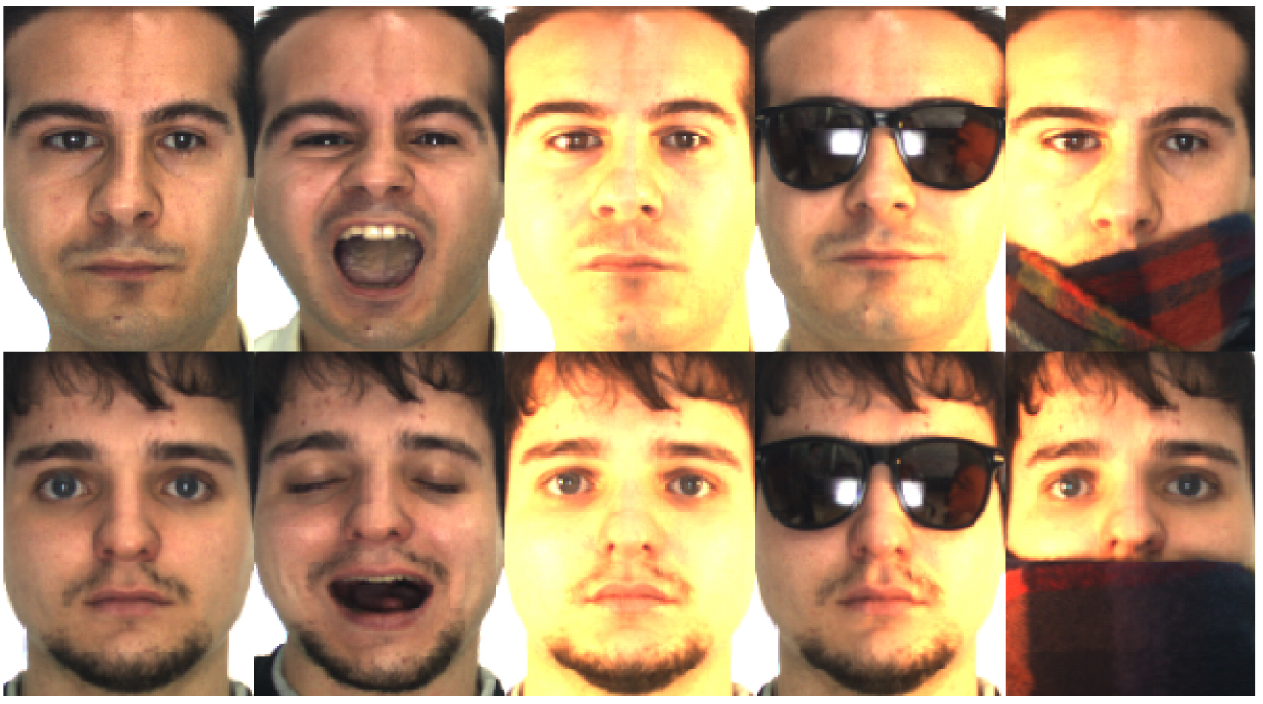}
  \caption{Example images from the AR database.}
  \label{fig:exam_ar}
\end{figure}

The AR database~\cite{martinez1998ar} consists of more than 4000 color images of 126 subjects (70 men and 56 women), these images have variations in facial expressions, illumination conditions and occlusions, example images from this database are shown in Fig.~\ref{fig:exam_ar}. Following the experimental settings in Ref.~\citenum{xu2019sparse}, in our experiments, we use a subset with only illumination and expression changes that contains 50 male subjects and 50 female subjects from the AR database. For each individual, 7 images from Session 1 are used as training samples, and the other 7 images from Session 2 as test samples. All the images are firstly cropped to 60$\times$43 pixels and projected to a subspace of dimensions 54, 120, and 300 by PCA. Experimental results are summarized in Table ~\ref{tab:tab_ar}, the balancing parameter $\lambda$ of CRNRC under dimensions 54, 120 and 300 are set to be 0.001, 0.01 and 0.001, respectively. One can see that our proposed CRNRC achieves the highest recognition accuracy under all the three reduced dimensions.

\begin{table}[]
\caption{Recognition accuracy (\%) of competing approaches on the AR database.}
\label{tab:tab_ar}
\centering
\begin{tabular}{llll}
\hline
Dim.     & 54                 & 120 & 300 \\ \hline
NSC~\cite{lee2005acquiring}                                    & 70.7             & 75.5   & 76.1   \\
SVM   & 81.6             & 89.3    & 91.6   \\
SRC~\cite{wright2009robust}   & 82.1            & 88.3    & 90.3   \\
CRC~\cite{zhang2011sparse}  & 80.3             & 90.1      & 93.8 \\
CROC~\cite{chi2013classification}   & 82.0             & 90.8     & 93.7  \\
ProCRC~\cite{cai2016probabilistic}  & 81.4             & 90.7     & 93.7  \\
NRC~\cite{xu2019sparse}   & 85.2             & \textbf{91.3}     & 93.3  \\
\textbf{CRNRC}                        & \textbf{85.7}    & \textbf{91.3}   & \textbf{94.0}    \\ \hline
\end{tabular}
\end{table}

\subsubsection{Experiments on the Extended Yale B Database}
\begin{figure}[htbp]
  \centering
  \includegraphics[trim={0mm 0mm 0mm 0mm},clip, width = .8\textwidth]{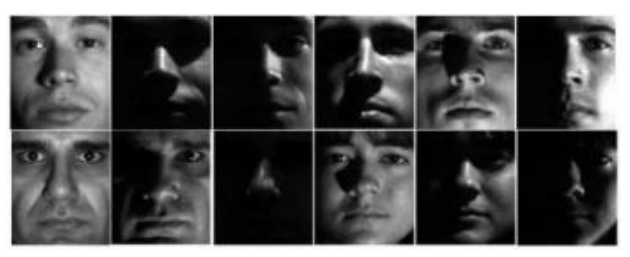}
  \caption{Example images from the Extended Yale B database.}
  \label{fig:exam_eyaleb}
\end{figure}

The Extended Yale B database~\cite{georghiades2001few} contains 2414 face images from 38 individuals, each having 59-64 images. These images have illumination variations, example images from this database are shown in Fig.~\ref{fig:exam_eyaleb}. The original images are of 192$\times$168 pixels. In our experiments, all the images are resized to 54$\times$48 pixels. 32 images per subject are randomly selected for training and the remaining for testing. The resized images are projected to a subspace of dimensions 84, 150, and 300 by PCA. Table~\ref{tab:tab_eyaleb} lists the classification accuracy of the competing approaches, the balancing parameter $\lambda$ of CRNRC under dimensions 84, 150 and 300 is set to be 0.001. It can be seen that CRNRC consistently outperforms its competing approaches in all cases.

\begin{table}[]
\caption{Recognition accuracy (\%) of competing approaches on the Extended Yale B database.}
\label{tab:tab_eyaleb}
\centering
\begin{tabular}{llll}
\hline
Dim.       & 84               & 150 & 300 \\ \hline
NSC~\cite{lee2005acquiring}           & 91.2             & 95.3   & 96.6   \\
SVM   & 93.4             & 95.8    & 96.9   \\
SRC~\cite{wright2009robust}         & 95.5            & 96.9    & 97.7   \\
CRC~\cite{zhang2011sparse}         & 95.0             & 96.3      & 97.8 \\
CROC~\cite{chi2013classification}   & 95.5             & 97.1     & 98.2  \\
ProCRC~\cite{cai2016probabilistic}   & 93.4             & 95.3     & 96.2  \\
NRC~\cite{xu2019sparse}         & 96.7             & 97.2     & \textbf{98.4}  \\
\textbf{CRNRC}                        & \textbf{96.8}    & \textbf{97.3}   & \textbf{98.4}    \\ \hline
\end{tabular}
\end{table}

\subsubsection{{\color{red}Experiments on the GT Database}}
\begin{figure}[htbp]
  \centering
  \includegraphics[trim={0mm 0mm 0mm 0mm},clip, width = .8\textwidth]{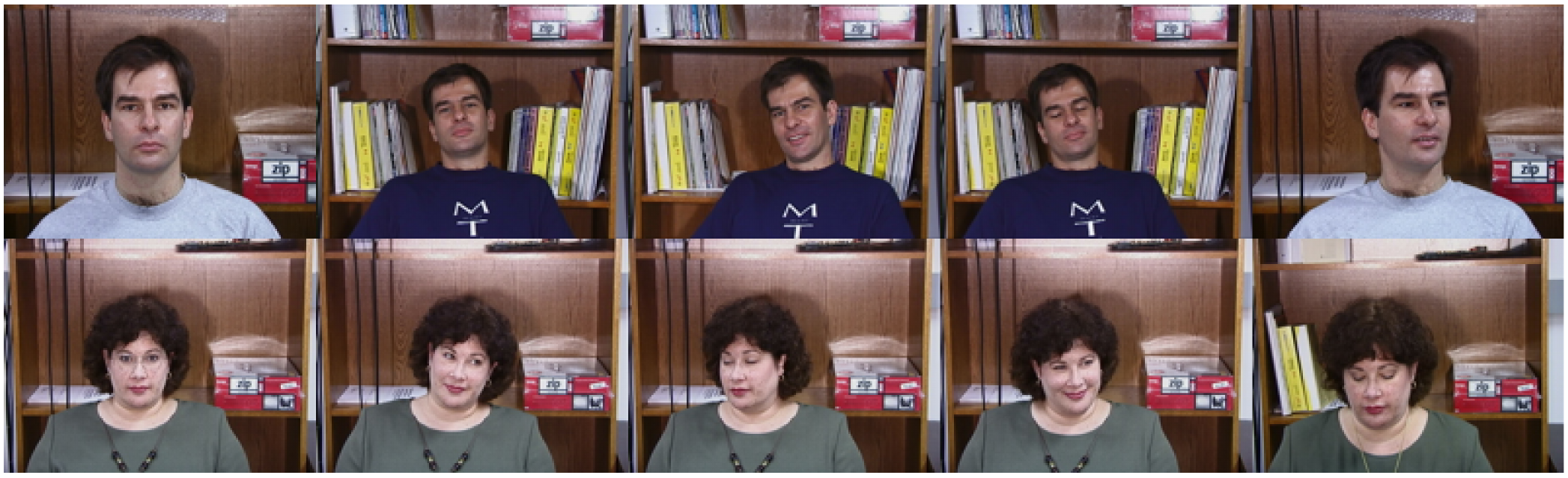}
  \caption{{\color{red}Example images from the GT database.}}
  \label{fig:exam_gt}
\end{figure}

{\color{red}The GT database~\cite{feng2016superimposed} contains 750 face images of 50 individuals, and each of them has 15 images taken at resolution of 480$\times$640 pixels. These images have variations in pose, expression, cluttered background, and illumination, some example images are depicted in Fig.~\ref{fig:exam_gt}. Apart from the comparison methods used on the AR and Extended Yale B databases, here we compare CRNRC with several prevailing deep architectures, \ie, GoogLeNet~\cite{szegedy2015going}, Inception-v3~\cite{szegedy2016rethinking}, Xception~\cite{chollet2017xception} and Inception-ResNet-v2~\cite{szegedy2017inception}. According to the experimental settings in Ref.~\citenum{mehdipour2016comprehensive}, we employ the pre-trained deep models to extract features, then these features are classified by the nearest neighbors with cosine metric. The layers of GoogLeNet, Inception-v3, Xception and Inception-ResNet-v2 for feature extraction are loss3-classifier, avg\_pool, avg\_pool, avg\_pool, respectively. The first eight images per subject are used as training samples, while the remaining as test samples. Experimental results are shown in Table~\ref{tab:tab_gt}, and the balancing parameter $\lambda$ of CRNRC is set as 0.001. We can observe that our proposed CRNRC outperforms the conventional RBCM as well as the deep architectures. Compared with Inception-ResNet-v2, CRNRC achieves a modest 0.3\% improvement.}

\begin{table}[]
\caption{{\color{red}Recognition accuracy (\%) of competing approaches on the GT database.}}
\label{tab:tab_gt}
\centering
\begin{tabular}{lllllll}
\hline
Methods  & {\color{red}NSC~\cite{lee2005acquiring}} & {\color{red}SVM}       & {\color{red}SRC~\cite{wright2009robust}}          & {\color{red}CRC~\cite{zhang2011sparse}}      & {\color{red}CROC~\cite{chi2013classification}}                & {\color{red}ProCRC~\cite{cai2016probabilistic}} \\ \hline
Accuracy &  {\color{red}90.2}   &  {\color{red}89.7}&   {\color{red}90.5}  &  {\color{red}91.0} &           {\color{red}92.0}   &  {\color{red}91.7} \\ \hline
Methods  & {\color{red}NRC~\cite{xu2019sparse}} & {\color{red}GoogLeNet~\cite{szegedy2015going}} & {\color{red}Inception-v3~\cite{szegedy2016rethinking}} & {\color{red}Xception~\cite{chollet2017xception}} & {\color{red}Inception-ResNet-v2~\cite{szegedy2017inception}} & {\color{red}CRNRC}  \\ \hline
Accuracy &  {\color{red}92.0}   &  {\color{red}90.8}   &   {\color{red}91.7} &         {\color{red}90.3} &  {\color{red}92.0}   &  {\color{red}92.3}  \\ \hline
\end{tabular}
\end{table}

\subsection{Handwritten Digit Classification}

\subsubsection{Experiments on the USPS Dataset}
The USPS dataset~\cite{hull1994database} is composed of 9298 images for digit numbers of ten classes, i.e. from 0 to 9. The training set and test set contain 7291 and 2007 images, respectively. All the images are resized into 16$\times$16 pixels. $N$ ($N$=50, 100, 200, 300) images per class from the training set are randomly selected for training and all the images in the test set for testing. Experiments are repeated for 10 times and the average results are recorded. Experimental results are shown in Table~\ref{tab:tab_usps}, the balancing parameter $\lambda$ of CRNRC is set to be 0.1 in all cases. We can observe that CRNRC achieves the best recognition result under all scenarios. With the increasing of the number of training images, recognition accuracy of all competing approaches improves steadily.

\begin{table}[]
\caption{Recognition accuracy (\%) of competing approaches on the USPS database.}
\label{tab:tab_usps}
\centering
\begin{tabular}{lllll}
\hline
$N$       &    50           & 100                 & 200&300 \\ \hline
NSC~\cite{lee2005acquiring}      & 91.2             & 92.2   & 92.8  & 92.8 \\
SVM   & 91.6             & 92.5    & 93.1 & 93.2  \\
SRC~\cite{wright2009robust}     &       89.1      &  91.2   & 92.9  & 93.8  \\
CRC~\cite{zhang2011sparse}    & 89.8             & 90.8      & 91.5 & 91.5\\
CROC~\cite{chi2013classification}   & 91.9      & 91.3     & 91.7 & 91.8 \\
ProCRC~\cite{cai2016probabilistic}     & 90.9     & 91.9     & 92.2 & 92.2 \\
NRC~\cite{xu2019sparse}   & 90.3             & 91.6     & 92.7 & 93.0 \\
\textbf{CRNRC}                        & \textbf{92.3}    & \textbf{93.6}   & \textbf{94.6} & \textbf{94.6}  \\ \hline
\end{tabular}
\end{table}

\subsubsection{Experiments on the MNIST Dataset}
The MNIST dataset~\cite{lecun1998gradient}  has 60,000 training samples and 10,000 testing samples for digit numbers from 0 to 9. The original images are of size 28$\times$28. We randomly selected $N$ ($N$=50, 100, 300, 500) samples per class from the training set for training, and utilize all the samples in the test set for testing. Experiments are repeated for 10 times and the average results are reported. Recognition accuracy of competing approaches is presented in Table~\ref{tab:tab_mnist}, the balancing parameter $\lambda$ of CRNRC is set to be 0.1 in all cases. One can see that when the number of training images per class is 50 and 100, CRNRC underperforms NSC. Unfortunately, with the increasing number of training images, the classification performance of NSC dramatically decreases. The reason behind this phenomenon lies in that NSC employs the class-specific training samples to represent the test sample, and the test sample can be well expressed by the training data when the number of training images per class is 50 and 100. Nevertheless, the training data matrix will be singular when the number of training images per class is 300 and 500, which will deteriorate the performance of NSC.

{\color{red}In addition to the recognition accuracy, here we present the F1 score of different methods on the MNIST dataset. The F1 score is the harmonic mean of precision and recall and therefore provides a single metric that summarizes the classifier performance in terms of both recall and precision. Table ~\ref{tab:fscore_mnist} summarizes the F1 score of competing approaches. It can be seen that with the increasing number of training samples per class, the F1 score of our proposed CRNRC is higher than that of the other approaches, which further demonstrates the effectiveness of our method.}
\begin{table}[]
\caption{Recognition accuracy (\%) of competing approaches on the MNIST database.}
\label{tab:tab_mnist}
\centering
\begin{tabular}{lllll}
\hline
$N$     &    50   &  100    &   300   &   500 \\ \hline
NSC~\cite{lee2005acquiring}       &     \textbf{91.6} & \textbf{92.7}   & 84.8  &  71.3\\
SVM   &  86.6   &  88.5   & 90.8  & 91.1  \\
SRC~\cite{wright2009robust}   &    82.4   &  86.7   & 91.2  & 92.7 \\
CRC~\cite{zhang2011sparse}  &     86.3   &   88.2    & 89.2  &  89.4\\
CROC~\cite{chi2013classification} &     91.1         &   92.5   & 88.7  & 87.2 \\
ProCRC~\cite{cai2016probabilistic}   &      86.6        &   89.5   & 92.5 & 93.5 \\
NRC~\cite{xu2019sparse}  &      86.1        &  88.5    & 90.7 & 91.7 \\
\textbf{CRNRC}  & 89.2    &  92.1  & \textbf{94.4}   & \textbf{95.1}  \\ \hline
\end{tabular}
\end{table}

\begin{table}[]
\caption{{\color{red}F1 score (\%) of competing approaches on the MNIST database.}}
\label{tab:fscore_mnist}
\centering
\begin{tabular}{lllll}
\hline
$N$     &    50   &  100    &   300   &   500 \\ \hline
NSC~\cite{lee2005acquiring}       &   {\color{red}91.5}   & {\color{red}92.7}  & {\color{red}84.9} & {\color{red}71.4} \\
SVM    &   {\color{red}86.4}   &  {\color{red}88.3}  & {\color{red}90.6}  & {\color{red}90.9} \\
SRC~\cite{wright2009robust}    &   {\color{red}82.0}   & {\color{red}86.5}  & {\color{red}91.1} & {\color{red}92.6} \\
CRC~\cite{zhang2011sparse}   &  {\color{red}86.0}    & {\color{red}88.0}  & {\color{red}89.0} & {\color{red}89.2} \\
CROC~\cite{chi2013classification}  &   {\color{red}91.0}   & {\color{red}92.4}  & {\color{red}88.5} & {\color{red}87.1} \\
ProCRC~\cite{cai2016probabilistic}   &  {\color{red}86.3}    & {\color{red}89.3}  & {\color{red}92.4} & {\color{red}93.4} \\
NRC~\cite{xu2019sparse}   &  {\color{red}85.8}    & {\color{red}88.2}  & {\color{red}90.6} & {\color{red}91.5} \\
\textbf{CRNRC}  & {\color{red}89.0} & {\color{red}92.0} & {\color{red}94.3} & {\color{red}95.0} \\ \hline
\end{tabular}
\end{table}

\subsection{Large Scale Pattern Classification}
To fully evaluate the performance of CRNRC, in this subsection we compare it with conventional RBCM and state-of-the-art approaches on four large scale datasets, \ie, Stanford 40 Actions dataset~\cite{yao2011human}, the Oxford 102 Flowers dataset\cite{nilsback2008automated}, the Aircraft dataset~\cite{maji2013fine} and the Cars dataset~\cite{krause20133d}. First we give a description of these datasets, then we evaluate our proposed CRNRC and its competing approaches on these datasets.

\begin{figure}[htbp]
  \centering
  \includegraphics[trim={0mm 0mm 0mm 0mm},clip, width = .8\textwidth]{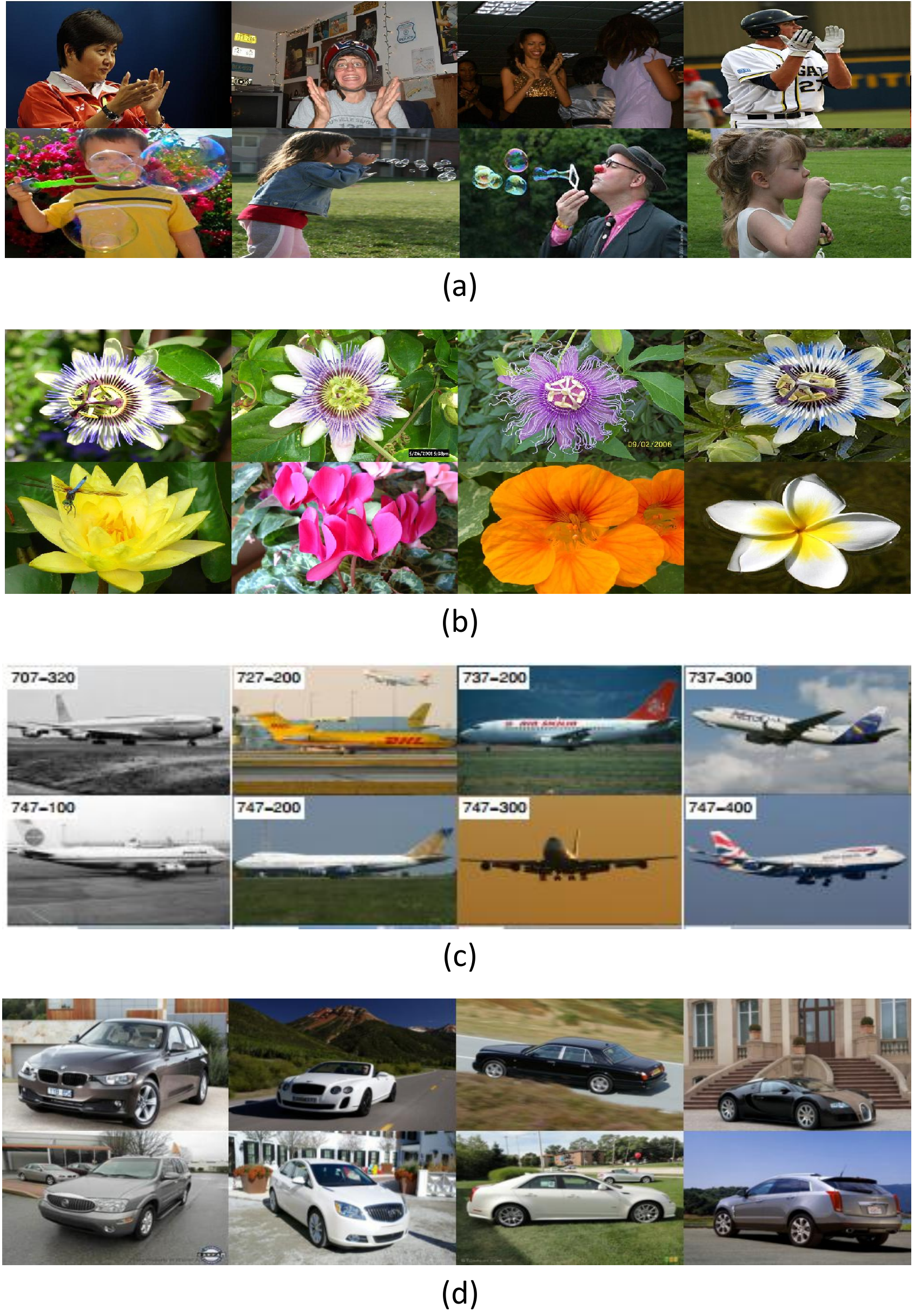}
  \caption{Example images from four large scale datasets. (a) Example images from the Stanford 40 dataset. (b) Example images from the Flower 102 dataset. (c) Example images from the Aircraft dataset. (d) Example images from the Cars dataset.}
  \label{fig:exam_fine}
\end{figure}

The Stanford 40 Actions dataset~\cite{yao2011human} is composed of 9532 images of humans performing 40 actions, such as reading book, throwing a frisbee and brushing teeth, example images from this dataset are shown in Fig.~\ref{fig:exam_fine} (a). Each action class has 180-300 images. Following the common training-testing split settings presented in Ref.~\citenum{xu2019sparse}, 100 images per class are randomly chosen for training and the remaining for testing.

The Oxford 102 Flowers dataset\cite{nilsback2008automated} contains 8189 images from 102 flower classes, example images from this dataset are shown in Fig.~\ref{fig:exam_fine} (b). The flowers chosen to be flower commonly occuring in the United Kingdom. Each class has 40-258 images, in which the images have large scale, pose and light variations.

The Aircraft dataset~\cite{maji2013fine} includes 10,000 images of aircraft spanning 100 aircraft models. The models appear at different scales, design structures, and appearances, making this dataset challenging for visual classification task, example images from this dataset are shown in Fig.~\ref{fig:exam_fine} (c). We use the same experimental settings as in Ref.~\citenum{xu2019sparse} to conduct our experiments.

The Cars dataset~\cite{krause20133d} has 16,185 images of 196 classes of cars, example images from this dataset are shown in Fig.~\ref{fig:exam_fine} (d). According to the standard split scheme, 8144 images are used as the training samples and the other 8041 images as the testing samples.

\subsubsection{Evaluation with Deep Features}
Following the experimental setting in Ref.~\citenum{xu2019sparse}, on the Stanford 40 Actions dataset and the Oxford 102 Flowers dataset, VGG-19~\cite{simonyan2014very} is employed to extract CNN features (referred to as VGG19 features), and the final feature dimension of each image is 4096 for these two datasets. For the Aircraft and Cars datasets, a VGG-16 network~\cite{simonyan2014very} is used to extract features and then these deep features are fed into CRNRC and its competing approaches. Recognition accuracy of competing approaches on the above four datasets are recorded in Table~\ref{tab:tab_large}, the balancing parameter $\lambda$ of CRNRC on the Stanford 14 dataset, Oxford 102 dataset, Aircraft dataset and Cars dataset are set to be 1e-3, 1e-5, 0.1 and 0.1, respectively. As we can see from Table~\ref{tab:tab_large}, CRNRC achieves the best result on the four datasets. Specifically, CRNRC makes an improvement of 0.2\%, 0.4\% and 0.6\% over NRC on the Stanford 40, Aircraft and Cars datasets, respectively.
\begin{table}[]
\caption{Recognition accuracy (\%) of competing approaches on the four fine-grained datasets.}
\label{tab:tab_large}
\centering
\begin{tabular}{lllll}
\hline
Methods    &    Stanford 40   & Flower 102    &   Aircraft  &  Cars \\ \hline
Softmax   &   77.2           &  87.3   & 85.6 &  88.7 \\
NSC~\cite{lee2005acquiring}       &   74.7           & 90.1   & 85.5  & 88.3 \\
SRC~\cite{wright2009robust}    &  78.7           &  93.2   & 86.1  &  89.2\\
CRC~\cite{zhang2011sparse}       &   78.2         &   93.0    &86.7 &  90.0\\
CROC~\cite{chi2013classification}   &     79.2    &   93.1   & 86.9 &  90.3\\
ProCRC~\cite{cai2016probabilistic}     &   80.9    &  94.8    & 86.8 & 90.1 \\
NRC~\cite{xu2019sparse}    &     81.1    &  \textbf{95.3}    &87.3  & 90.7 \\
\textbf{CRNRC}   &   \textbf{81.3}  &  \textbf{95.3}  & \textbf{87.7}  &  \textbf{91.1} \\ \hline
\end{tabular}
\end{table}

\subsubsection{Comparison with SOTA Methods}
In this section, we evaluate CRNRC with SOTA methods. As mentioned earlier, on the Stanford 40 Actions and Oxford 102 Flowers datasets, VGG19 features are used in NRC and CRNRC, while on the Aircraft and Cars datasets, VGG16 features are employed by NRC and CRNRC. It should be noted that the compared CNN based approaches exploit more sophisticated network architectures or features than our utilized VGG19 features.

On the Stanford 40 Actions dataset, AlexNet network~\cite{krizhevsky2012imagenet}, VGG19 network~\cite{simonyan2014very}, EPM~\cite{sharma2013expanded}, and ASPD~\cite{khan2015recognizing} are used for comparison. EPM and ASPD are two leading approaches on action recognition for still images. The classification accuracy is shown in Table~\ref{tab:tab_stan40}. One can see that CRNRC achieves the highest classification accuracy and outperforms NRC by 0.2\%.

On the Flower 102 dataset, AlexNet network~\cite{krizhevsky2012imagenet}, VGG19 network~\cite{simonyan2014very}, GMP~\cite{murray2014generalized}, OverFeat~\cite{sharif2014cnn}, and NAC~\cite{simon2015neural} are employed for comparison. Classification results are presented in Table~\ref{tab:tab_flowers}. We can observe that CRNRC has the same classification accuracy as NRC and NAC, and CRNRC achieves performance gains of 2.2\% over VGG19.

For the Aircraft and Cars datasets, VGG16 network~\cite{simonyan2014very}, Symbiotic~\cite{chai2013symbiotic}, FV-FGC~\cite{gosselin2014revisiting}, and B-CNN method~\cite{lin2015bilinear} are used for comparison, classification accuracy of competing approaches is summarized in Table~\ref{tab:tab_air_car}. Once again, CRNRC exhibits the best recognition performance, and it makes an improvement of 0.4\% and 0.4\% over NRC on the Aircraft dataset and Cars dataset, respectively.

\begin{table}[]
\caption{Recognition accuracy (\%) of competing approaches on the Stanford 40 actions dataset.}
\label{tab:tab_stan40}
\centering
\begin{tabular}{lllllll}
\hline
Methods    &    AlexNet~\cite{krizhevsky2012imagenet}   &EPM~\cite{sharma2013expanded}    &   ASPD~\cite{khan2015recognizing}   &  VGG19~\cite{simonyan2014very} &  NRC~\cite{xu2019sparse} & CRNRC  \\ \hline
Accuracy   &   68.6  &  72.3  &  75.4 &  77.2 & 81.1 & \textbf{81.3} \\ \hline
\end{tabular}
\end{table}

\begin{table}[]
\caption{Recognition accuracy (\%) of competing approaches on the Flower 102 dataset.}
\label{tab:tab_flowers}
\centering
\begin{tabular}{llllllll}
\hline
Methods    &    GMP~\cite{murray2014generalized}   & OverFeat~\cite{sharif2014cnn}    &   AlexNet~\cite{krizhevsky2012imagenet}  &  VGG19~\cite{simonyan2014very} &  NAC~\cite{simon2015neural} & NRC~\cite{xu2019sparse} & CRNRC  \\ \hline
Accuracy   &   84.6  &  86.8  &  90.4 &  93.1 & \textbf{95.3} & \textbf{95.3} & \textbf{95.3} \\ \hline
\end{tabular}
\end{table}

\begin{table}[]
\caption{Recognition accuracy (\%) of competing approaches on the Aircraft and Cars datasets.}
\label{tab:tab_air_car}
\centering
\begin{tabular}{lllllll}
\hline
Datasets    &    VGG16~\cite{simonyan2014very}   & Symbiotic~\cite{chai2013symbiotic}    &   FV-FGC~\cite{gosselin2014revisiting}  &  B-CNN~\cite{lin2015bilinear} &  NRC~\cite{xu2019sparse} & CRNRC  \\ \hline
Aircraft    &     85.6  &  72.5    & 80.7  & 84.1 & 87.3 & \textbf{87.7} \\
Cars   &   88.7  &  78.0  &  82.7 &  90.6 & 90.7 & \textbf{91.1} \\ \hline
\end{tabular}
\end{table}

\subsection{Parameter Sensitiveness Analysis}
\begin{figure}[htbp]
  \centering
  \includegraphics[trim={0mm 0mm 0mm 0mm},clip, width = .8\textwidth]{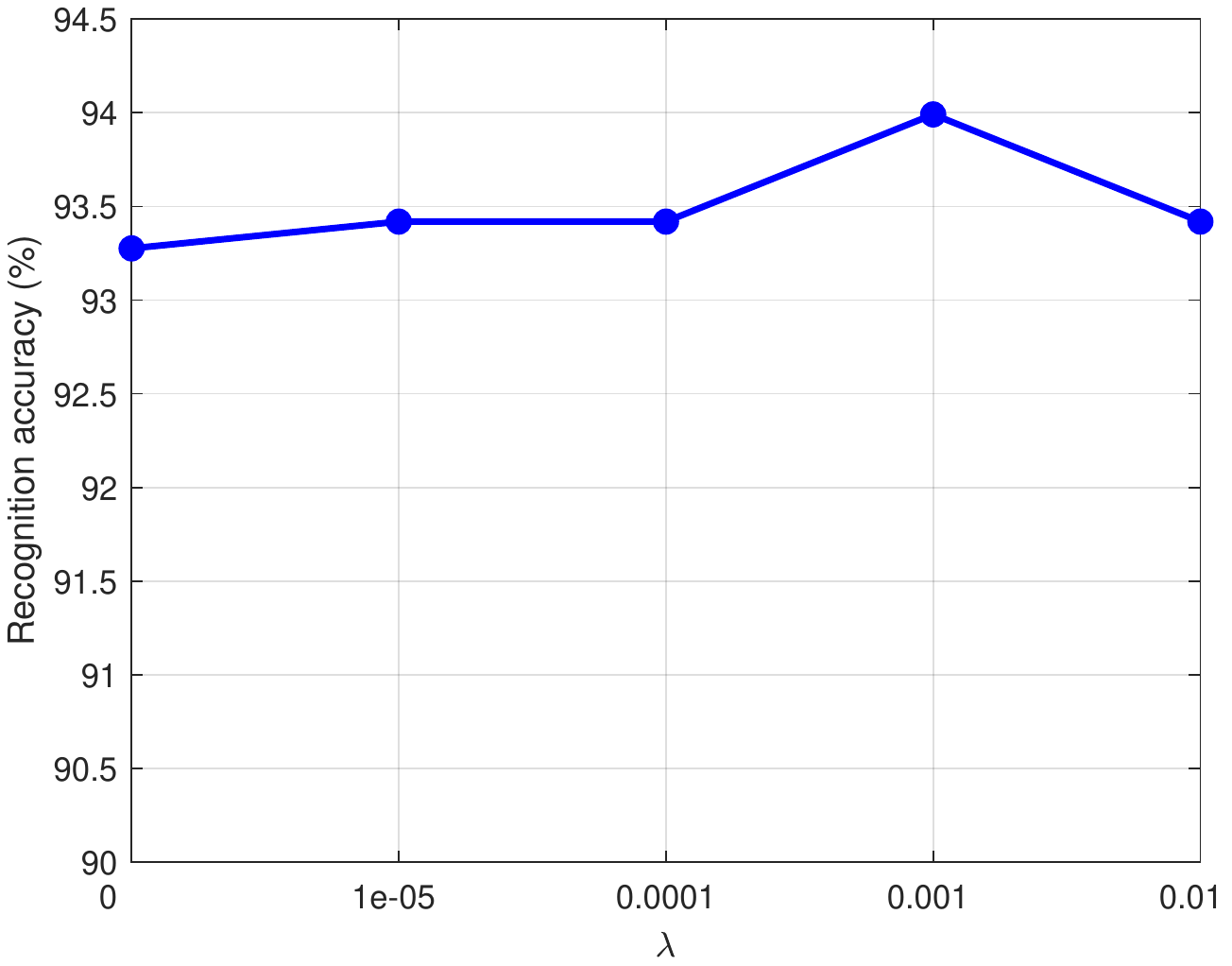}
  \caption{Classification accuracy (\%) of CRNRC with varying parameter $\lambda$ on the AR database.}
  \label{fig:param_lambda}
\end{figure}

To examine how the balancing parameter $\lambda$ influences the performance of CRNRC, we conduct experiments on the AR database. Experimental setting is the same as in Section \ref{sec:4_1_1} and the dimension of reduced samples is 300. Fig. \ref{fig:param_lambda} plots the recognition accuracy with varying $\lambda$. We can see that in quite a wide range of $\lambda$, CRNRC performs stable. Specifically, when $\lambda$ increases from 0 to 0.001, the classification accuracy of CRNRC also increases steadily. When $\lambda$ increases from 0.001 to 0.01, the performance of CRNRC drops a little. Larger value of $\lambda$ means that CRNRC will emphasize more on the class-specific residual constraint, which would undermine the collaborative mechanism of all training samples in representing the test sample. Therefore, we set a relatively small value for $\lambda$ in our experiments.

\section{Conclusions}
\label{sect:sect_5}
In this paper, we presented a class-specific residual constraint non-negative representation (CRNR) for pattern classification. Through this class-specific residual constraint, {\color{red}training samples from different classes are encouraged to competitively express the test sample.} The proposed NRCR is solved by the ADMM technique, and each subproblem has closed-form solution. Based on CRNR, we develop a CRNR based classifier (CRNRC). Experimental results on face databases, handwritten digit datasets and large-scale datasets validate that our proposed CRNRC outperforms NRC and several conventional RBCM, like SRC, CRC, CROC, and ProCRC. In this paper, we did not explicitly consider the situation that both the training and test samples are contaminated due to occlusion or corruption, thus in future, we will extend CRNRC to tackle the above scenarios.

\acknowledgments 
This work was supported by the National Natural Science Foundation of China (Grant No. 61672265, U1836218), the 111 Project of Ministry of Education of China (Grant No. B12018), the Postgraduate Research \& Practice Innovation Program of Jiangsu Province under Grant No. KYLX\_1123, the Overseas Studies Program for Postgraduates of Jiangnan University and the China Scholarship Council (CSC, No.201706790096). 


\bibliography{report}   

\begin{thebibliography}{10}

\bibitem{song2019collaborative}
X.~Song, Y.~Chen, Z.-H. Feng, {\em et~al.}, ``Collaborative representation
  based face classification exploiting block weighted lbp and analysis
  dictionary learning,'' {\em Pattern Recognition} {\bf 88}, 127--138  (2019).

\bibitem{jia2019collaborative}
S.~Jia, X.~Deng, J.~Zhu, {\em et~al.}, ``Collaborative representation-based
  multiscale superpixel fusion for hyperspectral image classification,'' {\em
  IEEE Transactions on Geoscience and Remote Sensing} {\bf 57}(10), 7770--7784
  (2019).

\bibitem{wright2009robust}
J.~Wright, A.~Y. Yang, A.~Ganesh, {\em et~al.}, ``Robust face recognition via
  sparse representation,'' {\em IEEE transactions on pattern analysis and
  machine intelligence} {\bf 31}(2), 210--227  (2009).

\bibitem{yu2009nonlinear}
K.~Yu, T.~Zhang, and Y.~Gong, ``Nonlinear learning using local coordinate
  coding,'' in {\em Advances in neural information processing systems},
  2223--2231  (2009).

\bibitem{wang2010locality}
J.~Wang, J.~Yang, K.~Yu, {\em et~al.}, ``Locality-constrained linear coding for
  image classification,'' in {\em 2010 IEEE computer society conference on
  computer vision and pattern recognition},  3360--3367, Citeseer  (2010).

\bibitem{lu2013face}
C.-Y. Lu, H.~Min, J.~Gui, {\em et~al.}, ``Face recognition via weighted sparse
  representation,'' {\em Journal of Visual Communication and Image
  Representation} {\bf 24}(2), 111--116  (2013).

\bibitem{li2010local}
C.-G. Li, J.~Guo, and H.-G. Zhang, ``Local sparse representation based
  classification,'' in {\em 2010 20th International Conference on Pattern
  Recognition},  649--652, IEEE  (2010).

\bibitem{ortiz2014face}
E.~G. Ortiz and B.~C. Becker, ``Face recognition for web-scale datasets,'' {\em
  Computer Vision and Image Understanding} {\bf 118}, 153--170  (2014).

\bibitem{zhang2011sparse}
L.~Zhang, M.~Yang, and X.~Feng, ``Sparse representation or collaborative
  representation: Which helps face recognition?,'' in {\em 2011 International
  conference on computer vision},  471--478, IEEE  (2011).

\bibitem{xu2016new}
Y.~Xu, Z.~Zhong, J.~Yang, {\em et~al.}, ``A new discriminative sparse
  representation method for robust face recognition via $ l\_ $\{$2$\}$ $
  regularization,'' {\em IEEE transactions on neural networks and learning
  systems} {\bf 28}(10), 2233--2242  (2016).

\bibitem{timofte2012weighted}
R.~Timofte and L.~Van~Gool, ``Weighted collaborative representation and
  classification of images,'' in {\em Proceedings of the 21st International
  Conference on Pattern Recognition (ICPR2012)},  1606--1610, IEEE  (2012).

\bibitem{chi2013classification}
Y.~Chi and F.~Porikli, ``Classification and boosting with multiple
  collaborative representations,'' {\em IEEE transactions on pattern analysis
  and machine intelligence} {\bf 36}(8), 1519--1531  (2013).

\bibitem{lee2005acquiring}
K.-C. Lee, J.~Ho, and D.~J. Kriegman, ``Acquiring linear subspaces for face
  recognition under variable lighting,'' {\em IEEE Transactions on Pattern
  Analysis \& Machine Intelligence} (5), 684--698  (2005).

\bibitem{cai2016probabilistic}
S.~Cai, L.~Zhang, W.~Zuo, {\em et~al.}, ``A probabilistic collaborative
  representation based approach for pattern classification,'' in {\em
  Proceedings of the IEEE conference on computer vision and pattern
  recognition},  2950--2959  (2016).

\bibitem{gou2019two}
J.~Gou, L.~Wang, B.~Hou, {\em et~al.}, ``Two-phase probabilistic collaborative
  representation-based classification,'' {\em Expert Systems with Applications}
  {\bf 133}, 9--20  (2019).

\bibitem{lee1999learning}
D.~D. Lee and H.~S. Seung, ``Learning the parts of objects by non-negative
  matrix factorization,'' {\em Nature} {\bf 401}(6755), 788  (1999).

\bibitem{xu2019sparse}
J.~Xu, W.~An, L.~Zhang, {\em et~al.}, ``Sparse, collaborative, or nonnegative
  representation: Which helps pattern classification?,'' {\em Pattern
  Recognition} {\bf 88}, 679--688  (2019).

\bibitem{boyd2011distributed}
S.~Boyd, N.~Parikh, E.~Chu, {\em et~al.}, ``Distributed optimization and
  statistical learning via the alternating direction method of multipliers,''
  {\em Foundations and Trends{\textregistered} in Machine learning} {\bf 3}(1),
  1--122  (2011).

\bibitem{martinez1998ar}
A.~M. Martinez, ``The ar face database,'' {\em CVC Technical Report24}
  (1998).

\bibitem{georghiades2001few}
A.~S. Georghiades, P.~N. Belhumeur, and D.~J. Kriegman, ``From few to many:
  Illumination cone models for face recognition under variable lighting and
  pose,'' {\em IEEE Transactions on Pattern Analysis \& Machine Intelligence}
  (6), 643--660  (2001).

\bibitem{feng2016superimposed}
Q.~Feng, C.~Yuan, J.-S. Pan, {\em et~al.}, ``Superimposed sparse parameter
  classifiers for face recognition,'' {\em IEEE transactions on cybernetics}
  {\bf 47}(2), 378--390  (2016).

\bibitem{hull1994database}
J.~J. Hull, ``A database for handwritten text recognition research,'' {\em IEEE
  Transactions on pattern analysis and machine intelligence} {\bf 16}(5),
  550--554  (1994).

\bibitem{lecun1998gradient}
Y.~LeCun, L.~Bottou, Y.~Bengio, {\em et~al.}, ``Gradient-based learning applied
  to document recognition,'' {\em Proceedings of the IEEE} {\bf 86}(11),
  2278--2324  (1998).

\bibitem{yao2011human}
B.~Yao, X.~Jiang, A.~Khosla, {\em et~al.}, ``Human action recognition by
  learning bases of action attributes and parts,'' in {\em 2011 International
  Conference on Computer Vision},  1331--1338, IEEE  (2011).

\bibitem{nilsback2008automated}
M.-E. Nilsback and A.~Zisserman, ``Automated flower classification over a large
  number of classes,'' in {\em 2008 Sixth Indian Conference on Computer Vision,
  Graphics \& Image Processing},  722--729, IEEE  (2008).

\bibitem{maji2013fine}
S.~Maji, E.~Rahtu, J.~Kannala, {\em et~al.}, ``Fine-grained visual
  classification of aircraft,'' {\em arXiv preprint arXiv:1306.5151}   (2013).

\bibitem{krause20133d}
J.~Krause, M.~Stark, J.~Deng, {\em et~al.}, ``3d object representations for
  fine-grained categorization,'' in {\em Proceedings of the IEEE International
  Conference on Computer Vision Workshops},  554--561  (2013).

\bibitem{szegedy2015going}
C.~Szegedy, W.~Liu, Y.~Jia, {\em et~al.}, ``Going deeper with convolutions,''
  in {\em Proceedings of the IEEE conference on computer vision and pattern
  recognition},  1--9  (2015).

\bibitem{szegedy2016rethinking}
C.~Szegedy, V.~Vanhoucke, S.~Ioffe, {\em et~al.}, ``Rethinking the inception
  architecture for computer vision,'' in {\em Proceedings of the IEEE
  conference on computer vision and pattern recognition},  2818--2826  (2016).

\bibitem{chollet2017xception}
F.~Chollet, ``Xception: Deep learning with depthwise separable convolutions,''
  in {\em Proceedings of the IEEE conference on computer vision and pattern
  recognition},  1251--1258  (2017).

\bibitem{szegedy2017inception}
C.~Szegedy, S.~Ioffe, V.~Vanhoucke, {\em et~al.}, ``Inception-v4,
  inception-resnet and the impact of residual connections on learning,'' in
  {\em Thirty-First AAAI Conference on Artificial Intelligence},   (2017).

\bibitem{mehdipour2016comprehensive}
M.~Mehdipour~Ghazi and H.~Kemal~Ekenel, ``A comprehensive analysis of deep
  learning based representation for face recognition,'' in {\em Proceedings of
  the IEEE conference on computer vision and pattern recognition workshops},
  34--41  (2016).

\bibitem{simonyan2014very}
K.~Simonyan and A.~Zisserman, ``Very deep convolutional networks for
  large-scale image recognition,'' {\em arXiv preprint arXiv:1409.1556}
  (2014).

\bibitem{krizhevsky2012imagenet}
A.~Krizhevsky, I.~Sutskever, and G.~E. Hinton, ``Imagenet classification with
  deep convolutional neural networks,'' in {\em Advances in neural information
  processing systems},  1097--1105  (2012).

\bibitem{sharma2013expanded}
G.~Sharma, F.~Jurie, and C.~Schmid, ``Expanded parts model for human attribute
  and action recognition in still images,'' in {\em proceedings of the IEEE
  Conference on Computer Vision and Pattern Recognition},  652--659  (2013).

\bibitem{khan2015recognizing}
F.~S. Khan, J.~Xu, J.~Van De~Weijer, {\em et~al.}, ``Recognizing actions
  through action-specific person detection,'' {\em IEEE transactions on image
  processing} {\bf 24}(11), 4422--4432  (2015).

\bibitem{murray2014generalized}
N.~Murray and F.~Perronnin, ``Generalized max pooling,'' in {\em Proceedings of
  the IEEE Conference on Computer Vision and Pattern Recognition},  2473--2480
  (2014).

\bibitem{sharif2014cnn}
A.~Sharif~Razavian, H.~Azizpour, J.~Sullivan, {\em et~al.}, ``Cnn features
  off-the-shelf: an astounding baseline for recognition,'' in {\em Proceedings
  of the IEEE conference on computer vision and pattern recognition workshops},
   806--813  (2014).

\bibitem{simon2015neural}
M.~Simon and E.~Rodner, ``Neural activation constellations: Unsupervised part
  model discovery with convolutional networks,'' in {\em Proceedings of the
  IEEE International Conference on Computer Vision},  1143--1151  (2015).

\bibitem{chai2013symbiotic}
Y.~Chai, V.~Lempitsky, and A.~Zisserman, ``Symbiotic segmentation and part
  localization for fine-grained categorization,'' in {\em Proceedings of the
  IEEE International Conference on Computer Vision},  321--328  (2013).

\bibitem{gosselin2014revisiting}
P.-H. Gosselin, N.~Murray, H.~J{\'e}gou, {\em et~al.}, ``Revisiting the fisher
  vector for fine-grained classification,'' {\em Pattern recognition letters}
  {\bf 49}, 92--98  (2014).

\bibitem{lin2015bilinear}
T.-Y. Lin, A.~RoyChowdhury, and S.~Maji, ``Bilinear cnn models for fine-grained
  visual recognition,'' in {\em Proceedings of the IEEE international
  conference on computer vision},  1449--1457  (2015).

\end{thebibliography}
\bibliographystyle{spiejour}   


\vspace{1ex}\noindent\textbf{He-Feng Yin} received his B.S. degree in School of Computer Science and Technology from Xuchang University, Xuchang, China, in 2011. Currently, he is a PhD candidate in School of IoT Engineering, Jiangnan University, Wuxi, China. He was a visiting PhD student at centre for vision, speech and signal processing (CVSSP), University of Surrey, under the supervision of Prof. Josef Kittler. His research interests include representation based classification methods, dictionary learning and low rank representation.

\vspace{1ex}\noindent\textbf{Xiao-Jun Wu} received the B.S. degree in mathematics from Nanjing Normal University, Nanjing, China, in 1991. He received the M.S. degree in 1996, and the Ph.D. degree in pattern recognition and intelligent systems in 2002, both from Nanjing University of Science and Technology, Nanjing, China. He joined Jiangnan University in 2006, where he is currently a Professor. He has published more than 200 papers in his fields of research. He was a visiting researcher in the Centre for Vision, Speech, and Signal Processing (CVSSP), University of Surrey, U.K., from 2003 to 2004. His current research interests include pattern recognition, computer vision, fuzzy systems, neural networks, and intelligent systems.

\listoffigures
\listoftables

\end{spacing}
\end{document}